\documentclass[conference]{IEEEtran}
\usepackage[T1]{fontenc}

\usepackage{cite}
\usepackage{amsmath,amssymb,amsfonts}
\usepackage{graphicx}
\usepackage{booktabs}
\usepackage{textcomp}
\usepackage{xcolor}
\usepackage{url}

\newcommand{\para}[1]{\par\noindent\textbf{#1}\ }
\newcommand{\dbetween}{\Delta_{\mathrm{between}}}
\newcommand{\swithin}{\bar{\sigma}_{\mathrm{within}}}

\begin{document}

\title{Denoised but Not Improved:\\What EEG Denoising Does to a Downstream Task}

\author{\IEEEauthorblockN{Xianglong Bao, Zhe Wang, Kewen Wang, Ming Jin\textsuperscript{*}}
\IEEEauthorblockA{\textit{School of Information and Communication Technology, Griffith University}, Brisbane, Australia \\
xianglong.bao@griffithuni.edu.au, \{zhe.wang, k.wang\}@griffith.edu.au, mingjinedu@gmail.com \\ \textsuperscript{*}Corresponding author}
}

\maketitle

\begin{abstract}
Scalp electroencephalography (EEG) is routinely denoised before it reaches a downstream model, and the denoising literature scores a method by how closely its output matches a clean reference waveform. That reference does not exist for real scalp recordings: it is synthesized by adding artifact templates to signals merely presumed clean, and even an exact separation would return cortical activity blurred by volume conduction rather than the activity itself. Fidelity to it is therefore an unvalidatable objective, and whether it predicts any downstream benefit has gone largely unexamined. We report an audit that examines it. Holding a seizure-detection pipeline fixed and varying only the preprocessing stage, we find that five standard denoising methods, spanning analytic, cross-channel and supervised deep families, all fail to improve on the unprocessed signal, and most degrade it. To say what such a stage must instead do, we formalize a label-only, classifier-free probe that reads it by the separability geometry it induces on task-relevant features, decomposing any change into a between-class distance and a within-class scatter term. On a transform that does improve the task, the probe attributes the entire gain to a roughly one-quarter reduction of within-class scatter, with class separation statistically unchanged. What a downstream model benefits from is therefore variance suppression on task-relevant structure, which waveform-fidelity metrics neither measure nor reward.
\end{abstract}


\section{Introduction}

Scalp electroencephalography (EEG) is a degraded observation of cortical activity. Volume conduction through the skull and scalp blurs the underlying neural sources~\cite{nunez2006electric}, and extracranial nuisance from the eyes, muscles and recording environment overlaps the neural content~\cite{uriguen2015eeg}. Because of this, denoising is a near-universal preprocessing stage in EEG pipelines: it consumes the raw recording and emits a signal of the same shape, which a downstream model then treats as its input. Classical denoisers decompose a multi-channel recording and discard the components judged artifactual~\cite{jung2000blind,hyvarinen2000independent,delorme2007enhanced}, shrink a wavelet expansion~\cite{donoho1994ideal}, or reject segments whose variance exceeds a calibration statistic~\cite{mullen2015real}. Deep denoisers replace these rules with a network trained to regress a clean recording from a corrupted one~\cite{zhang2021eegdenoisenet,chuang2025art,huang2025eegdfus,choi2025targetedadversarialdenoisingautoencoders}.

Almost all of them are developed and reported against the same objective: \emph{waveform fidelity}, the agreement between the denoised output and a clean reference, measured by relative root mean squared error or a correlation coefficient. The difficulty is that no such reference exists for a real scalp recording. Deep denoisers obtain one by synthesis, adding artifact templates to recordings that are themselves only presumed clean, so the network learns the corruption process its designer wrote down rather than the one the scalp imposes. Classical denoisers need no templates, but the separation they perform is blind~\cite{jung2000blind} and its solution unobservable, which leaves the labelling of components to a heuristic that no ground truth can check. Beneath both lies a more basic obstacle: even an exact separation would return cortical activity blurred by volume conduction~\cite{nunez2006electric,kayser2015surface} rather than the activity itself, so there is no clean scalp target to be verified against even in principle.

An objective that cannot be validated is not automatically the wrong one, but it does place the burden on some external check. The natural check is the downstream task the denoising is meant to serve, and it is precisely the one the denoising literature reports least often: methods are compared on fidelity to synthetic references and then deployed ahead of task models on the assumption that the two move together. This paper examines that assumption.

We make two contributions. \textbf{(i)} We report a controlled audit. Holding a clinical seizure-detection pipeline entirely fixed and varying only the preprocessing stage, we evaluate five standard denoising methods spanning the analytic, cross-channel and supervised-deep families. None improves on the unprocessed signal; each falls below it on binary F1 and on the Matthews correlation coefficient, and the supervised ones degrade the ranking of ictal against interictal windows outright. Denoising the scalp signal by these means is therefore not sufficient to help the model that consumes it. \textbf{(ii)} We formalize the label-only probe of~\cite{bao2026ieeg}, which says what a preprocessing stage \emph{does} in the absence of any clean reference. It measures separability between the task classes in a space of clinical morphology descriptors and splits any change into a between-class distance term and a within-class scatter term, distinguishing a stage that suppresses nuisance variance from one that amplifies class structure. On a transform that \emph{does} improve the task, the entire gain is attributable to a roughly one-quarter reduction of within-class scatter, with class separation statistically unchanged. What a downstream model benefits from, on this evidence, is variance suppression on task-relevant structure, and that is not what a waveform-fidelity score measures.

\section{Why Waveform Fidelity Cannot Be Validated}
\label{sec:why}

Write the neural activity of interest as a latent process $s(t)$. A scalp electrode observes
\begin{equation}
  x(t) \;=\; A\,s(t) \;+\; \zeta(t),
  \label{eq:obs}
\end{equation}
where $A$ implements volume conduction through the tissue layers between cortex and scalp~\cite{nunez2006electric} and $\zeta$ aggregates the extracranial nuisance contributed by the eyes, the muscles and the recording environment~\cite{uriguen2015eeg}. A denoiser is a map $h$ that returns $\tilde{x} = h(x)$ of the same shape, intended to be closer to the neural content than $x$ was.

Two properties of Eq.~\eqref{eq:obs} obstruct any fidelity-based evaluation of $h$. First, the decomposition is unobservable: $\zeta$ cannot be explicitly separated from $As$ given only $x$, so recovering the first term is a blind source separation problem whose solution is never revealed~\cite{jung2000blind}, and the extent to which a given $h$ has achieved it cannot be measured on a real recording. Second, and more fundamentally, the target itself is unattainable: $A$ is non-injective, so even an oracle that removed $\zeta$ exactly would leave $As$, a blurred image of $s$ rather than $s$~\cite{nunez2006electric,kayser2015surface}. There is no operation on a scalp recording whose correct output is the neural process, and therefore no reference against which fidelity could be defined.

Benchmarks resolve this by constructing a reference. A recording presumed clean is designated $x_{\mathrm{clean}}$, an artifact template is added at a controlled signal-to-noise ratio to obtain $x_{\mathrm{noisy}}$, and $h$ is fitted and scored by the agreement between $h(x_{\mathrm{noisy}})$ and $x_{\mathrm{clean}}$~\cite{zhang2021eegdenoisenet}. This makes the objective well-posed but relocates the question rather than answering it. The score now measures how well $h$ inverts a corruption model that was written down, and $x_{\mathrm{clean}}$ inherits whatever contamination the presumption missed, while remaining, by the argument above, a blurred observation in any case. A denoiser can therefore attain a high fidelity score without having become more useful to anything that consumes its output, and nothing in the fidelity score reveals whether it has.

This is not an argument that fidelity-trained denoisers are worthless, but that fidelity cannot be the terminal criterion. Sections~\ref{sec:probe} and~\ref{sec:audit} supply two external readouts: the downstream task, and the separability geometry the preprocessing induces.

\section{A Label-Only Separability Probe}
\label{sec:probe}

The downstream task is the criterion that matters, but a task score alone says only \emph{whether} a preprocessing stage helped, not \emph{what} it did, and it entangles that stage with the capacity and inductive bias of whichever model was trained on top. We therefore want a second readout that is independent of any trained model, requires no clean reference, and reports the effect in interpretable terms. The following probe, adapted from~\cite{bao2026ieeg}, meets these constraints; it needs only the class labels the task already provides.

\para{Feature space.} We map each signal window to six morphology descriptors long used in clinical EEG analysis: line length~\cite{esteller2001line}, which is the canonical seizure biomarker; the three Hjorth parameters, activity, mobility and complexity~\cite{hjorth1970eeg}; kurtosis~\cite{alam2013detection}; and the zero-crossing rate~\cite{pyrzowski2021zero}. These are fixed, interpretable and cheap, and none is learned, so the probe cannot adapt itself to flatter a particular method.

\para{Separability index.} Let $\mu_{\mathrm{ict}}, \mu_{\mathrm{int}}$ be the class centroids in this feature space and $\sigma^2_{\mathrm{ict}}, \sigma^2_{\mathrm{int}}$ their within-class spreads. We measure class separability by the standardized mean difference, Cohen's $d$~\cite{cohen1988statistical}, equivalently the sensitivity index of signal-detection theory~\cite{green1966signal},
\begin{equation}
  d \;=\; \frac{\lVert \mu_{\mathrm{ict}} - \mu_{\mathrm{int}} \rVert}{\sqrt{\tfrac{1}{2}\left(\sigma^2_{\mathrm{ict}} + \sigma^2_{\mathrm{int}}\right)}}
  \;=\; \frac{\dbetween}{\swithin}.
  \label{eq:d}
\end{equation}
The index is computed per subject, before and after processing, and compared by a paired test across subjects, so each subject serves as their own control and between-subject variability does not enter.

\para{Decomposition.} The value of Eq.~\eqref{eq:d} is that it is a ratio of two separately measurable quantities, so a stage that raises it can be asked \emph{how}. Only two routes exist. It may increase the numerator $\dbetween$, pushing the class centroids apart; we call this \emph{pattern separation}, and it is what an operation that amplifies class-specific structure would do. Or it may decrease the denominator $\swithin$, tightening each class about its centroid; we call this \emph{noise suppression}, and it is what an operation that removes variation uncorrelated with the label would do. Reporting the two terms separately turns an opaque improvement into a statement about mechanism. The distinction matters clinically: a stage that amplifies whatever it takes to be pathological is a different object, with different failure modes, from one that quiets everything not carrying label information.

\para{Properties.} The probe needs no clean reference, since it never compares the output to a target; no trained classifier, since it reads the feature space directly; and no fitting of its own. It is therefore cheap enough to run for every candidate method, cannot be gamed by one that merely suits a particular downstream architecture, and is a companion to the task score rather than a substitute.

\section{Audit}
\label{sec:audit}

\subsection{Setup}

\para{Pipeline.} All experiments use one pipeline, that of~\cite{bao2026ieeg}, with the preprocessing stage as the only free component. The task is seizure detection on CHB-MIT~\cite{shoeb2010chbmit}, a pediatric scalp database from Children's Hospital Boston, from which we retain 22 patients, excluding one whose channel data are unreadable and one repeat recording~\cite{thuwajit2022eegwavenet}. Recordings are taken in the 18-channel longitudinal bipolar montage at 256\,Hz and segmented into 4\,s windows at 1\,s stride, a window being labelled ictal if it overlaps a clinician-annotated seizure by at least 50\%. Evaluation is subject-dependent leave-one-recording-out cross-validation, training on the remaining recordings of the same patient, so no window from the held-out recording can leak into training. This yields a per-patient score distribution, which supports the cross-patient paired tests used throughout. The downstream model is EEGWaveNet~\cite{thuwajit2022eegwavenet}, a convolutional seizure detector, trained with its original protocol and never informed of how its input was preprocessed.

\begin{table}[t]
\centering
\caption{Downstream seizure detection on CHB-MIT with EEGWaveNet as the host. Every row differs only in the preprocessing occupying one fixed position in the pipeline; the host, its training protocol, the cross-validation split and the metrics are identical throughout. Scores are means over 22 patients, as reported in~\cite{bao2026ieeg}.}
\label{tab:audit}
\setlength{\tabcolsep}{2.4pt}
\footnotesize
\begin{tabular}{lcccccc}
\toprule
Preprocessing & BF1 & Sn & Pr & BAcc & MCC & AUROC \\
\midrule
None (raw signal) & 0.681 & 0.717 & 0.706 & 0.852 & 0.688 & 0.955 \\
\midrule
Wavelet thresholding & 0.657 & 0.733 & 0.654 & 0.852 & 0.662 & 0.955 \\
ASR                  & 0.667 & 0.732 & 0.690 & 0.848 & 0.674 & 0.953 \\
ICA                  & 0.661 & 0.697 & 0.705 & 0.841 & 0.671 & 0.956 \\
EEGdenoiseNet (EOG)  & 0.604 & 0.651 & 0.630 & 0.817 & 0.611 & 0.945 \\
EEGdenoiseNet (EMG)  & 0.626 & 0.661 & 0.650 & 0.823 & 0.631 & 0.939 \\
\midrule
Positive control & \textbf{0.749} & \textbf{0.776} & \textbf{0.774} & \textbf{0.882} & \textbf{0.754} & \textbf{0.972} \\
\bottomrule
\end{tabular}
\end{table}

\para{Metrics.} Seizure data is heavily imbalanced, with interictal background windows far outnumbering ictal ones, so conventional accuracy approaches 100\% even for a detector that mostly predicts background. We therefore report six metrics that stay informative under imbalance and are not dominated by the majority class: binary F1 on the ictal class (BF1), sensitivity (Sn), precision (Pr), balanced accuracy (BAcc), the Matthews correlation coefficient (MCC), and AUROC.

\para{Denoisers under audit.} We evaluate five denoisers spanning the families into which EEG denoising falls. \emph{Wavelet thresholding}~\cite{donoho1994ideal} expands each channel over a wavelet basis, estimates the noise level from the finest scale, and shrinks coefficients below a universal threshold; its dictionary is analytic, not learned. \emph{Artifact subspace reconstruction} (ASR)~\cite{mullen2015real} calibrates on the low-variance portion of each recording and reconstructs the components whose variance exceeds a threshold. \emph{Independent component analysis} (ICA)~\cite{hyvarinen2000independent} decomposes each recording into maximally independent components and discards those whose kurtosis identifies them as artifactual~\cite{delorme2007enhanced}. These three are unsupervised. \emph{EEGdenoiseNet}~\cite{zhang2021eegdenoisenet} is supervised; following the benchmark's own protocol, which fits a separate network per artifact type, we train its Novel CNN on the benchmark's paired data at controlled signal-to-noise ratios, once for ocular (EOG) and once for myogenic (EMG) contamination, giving two audited denoisers.

\para{A positive control.} An audit that only reports failures cannot distinguish ``nothing helps this task'' from ``these denoisers do not''. We therefore include one transform that is known to improve this pipeline~\cite{bao2026ieeg}, placed at the same point and returning a signal of the same shape, and use it in Section~\ref{sec:helpful} to ask what a helpful stage does. It is deliberately not a denoiser: it is never trained to remove any artifact and optimizes no fidelity criterion. It is a convolutional encoder--decoder with a residual vector-quantized bottleneck, of the kind developed for neural audio codecs~\cite{defossez2022encodec} and since applied to brain signals~\cite{carzaniga2025braincodec}, quantizing each latent step against four residual codebooks of 256 entries. Two properties matter here. First, it is fitted on \emph{intracranial} EEG, from the SWEC database of 16 presurgical epilepsy patients~\cite{burrello2019swec}, a modality that observes cortical activity with far less contamination because the electrode bypasses the dominant scalp pathways~\cite{ball2009signal,parvizi2018promises}; it therefore never sees a scalp recording during fitting and needs no clean scalp target. Second, it is then \emph{frozen}, applied channel-wise and never tuned to the task, so it is a fixed transformation exactly as the five audited denoisers are.

\subsection{No Audited Denoiser Improves the Task}

Table~\ref{tab:audit} reports the audit. Not one of the five improves on the raw signal. Each falls below it on binary F1 and on MCC, the two metrics least forgiving of a detector that trades one error type for another, and the shortfall is not marginal for the supervised pair, which give up 0.077 and 0.055 binary F1 against doing nothing at all.

The unsupervised and supervised failures have different shapes, and the AUROC column separates them. Wavelet thresholding, ASR and ICA hold AUROC within 0.002 of raw, so they barely alter how the model \emph{ranks} ictal against interictal windows; what they perturb is where the operating point falls, legible in the precision column, where wavelet thresholding gives up 0.052 precision to buy 0.016 sensitivity. These three are close to inert with respect to the task: they remove variance the detector was not using, at the cost of some it was. The supervised variants are worse, lowering AUROC by 0.010 and 0.016 and so degrading the ranking itself, which no threshold can recover. This is what Section~\ref{sec:why} anticipates: a network fitted to invert a written-down corruption model has been optimized for a criterion that is not merely unvalidatable but, here, actively misaligned, since the structure it learned to remove as ocular or myogenic contamination overlaps structure the detector was using.

The positive control, by contrast, improves every metric, by 0.068 binary F1 and 0.066 MCC, and raises AUROC as well, so the gain survives the removal of any threshold effect. The task is therefore not one on which preprocessing cannot help. What separates this transform from the five denoisers is not that it removes artifacts better: it sees no scalp data at all and optimizes no fidelity criterion. The next subsection asks what it does instead.

\subsection{What a Helpful Stage Actually Does}
\label{sec:helpful}

\begin{figure}[t]
\centering
\includegraphics[width=\linewidth]{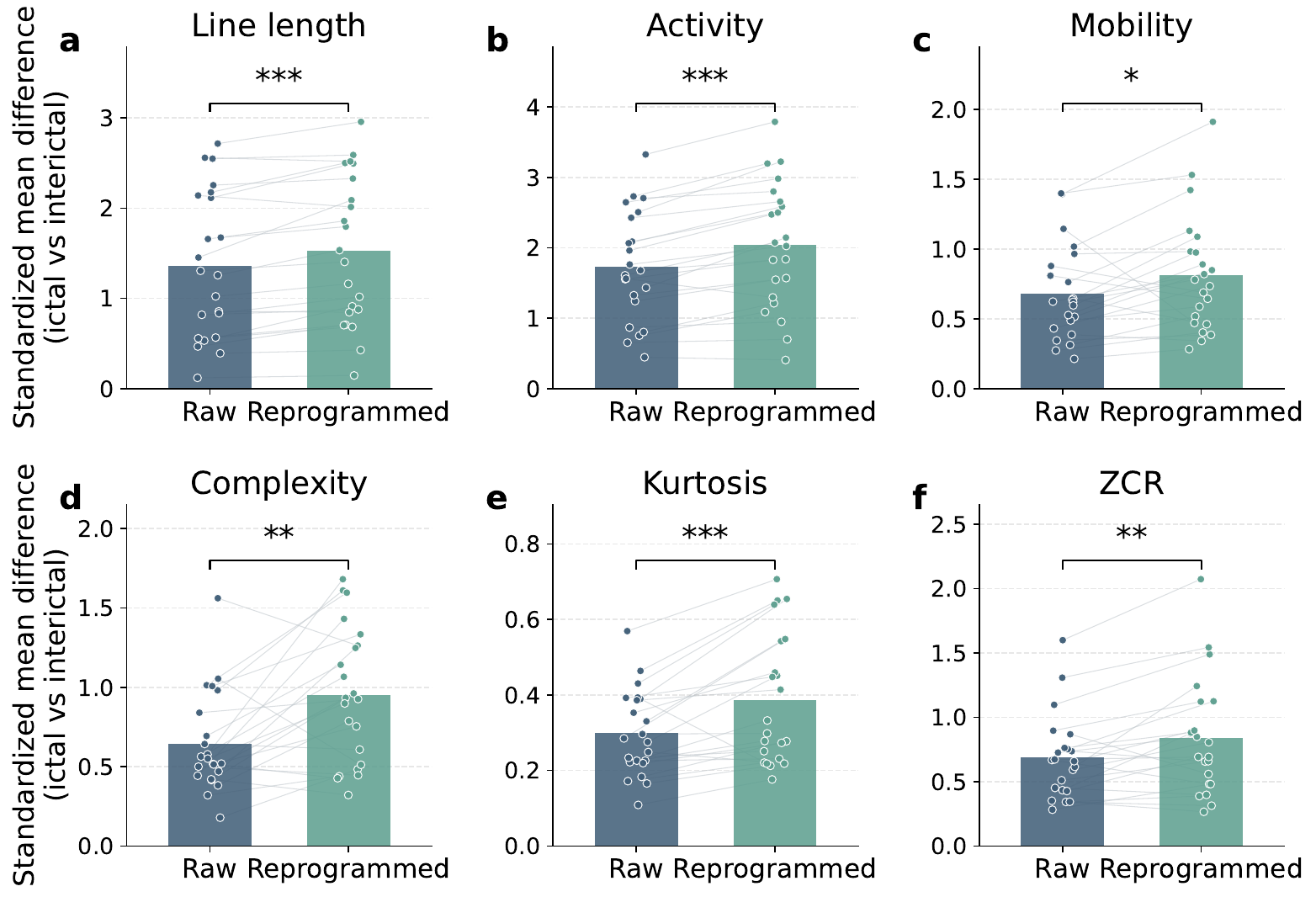}
\caption{Per-descriptor separability before and after the positive control. Each panel is one morphology descriptor, each dot a per-patient standardized mean difference between ictal and interictal windows, joined across conditions (\emph{Reprogrammed} denotes the positive-control output). All six improve significantly (${}^{*}p<0.05$, ${}^{**}p<0.01$, ${}^{***}p<0.001$), so the joint gain is broad rather than carried by any single statistic. Adapted from~\cite{bao2026ieeg}.}
\label{fig:perfeature}
\end{figure}

We run the probe of Section~\ref{sec:probe} on the positive control, per patient, over the same 22 patients and the same windows. In the joint six-descriptor space, separability rises from $d = 0.93$ on the raw signal to $d = 1.26$ after processing, a 35\% increase, significant across patients at $p < 10^{-6}$ by a Wilcoxon signed-rank test. The effect is not carried by one descriptor: as Fig.~\ref{fig:perfeature} shows, all six improve significantly on their own, with line length, Hjorth activity and kurtosis at $p < 0.001$, Hjorth complexity and the zero-crossing rate at $p < 0.01$, and Hjorth mobility at $p < 0.05$.

\begin{figure}[t]
\centering
\includegraphics[width=0.76\linewidth]{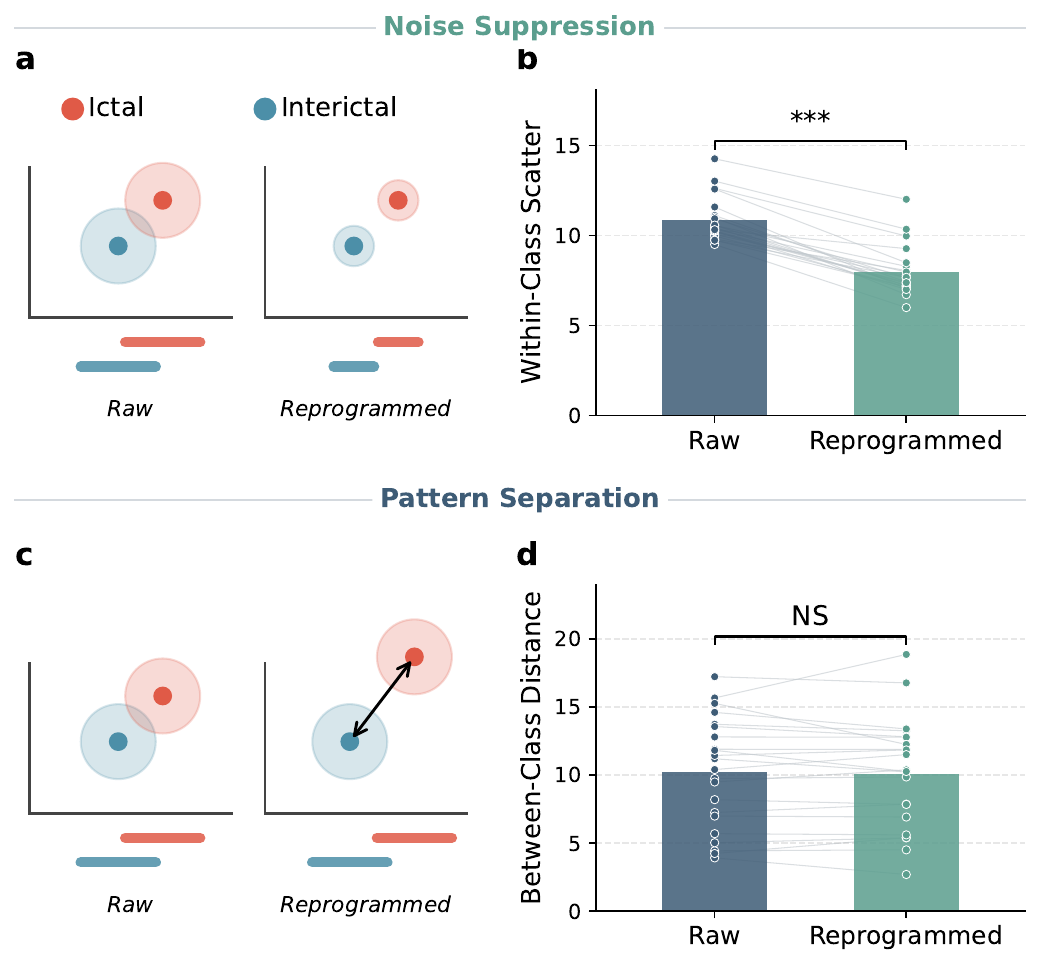}
\caption{Decomposing the separability change under the positive control, over 22 patients. \textbf{(a,\,b)} Noise suppression: within-class scatter $\swithin$ falls by about a quarter. \textbf{(c,\,d)} Pattern separation: the between-class centroid distance $\dbetween$ is unchanged. Axis labels: \emph{Raw} is the unprocessed signal, \emph{Reprogrammed} the positive-control output. Dots are individual patients, joined between conditions; ${}^{***}p<0.001$, NS = not significant, cross-patient Wilcoxon signed-rank. Adapted from~\cite{bao2026ieeg}.}
\label{fig:decomp}
\end{figure}

Decomposing the index answers the question the task score cannot. Fig.~\ref{fig:decomp} reports the two terms of Eq.~\eqref{eq:d} separately. The between-class centroid distance $\dbetween$ is statistically unchanged: ictal and interictal activity are no further apart after processing than before. The within-class scatter $\swithin$ falls by roughly a quarter, significant at $p < 0.001$, and this term alone accounts for the entire rise in $d$, consistently with the ratio of Eq.~\eqref{eq:d}. The transform is thus a pure noise suppressor in the sense of Section~\ref{sec:probe} and not a pattern amplifier. It does not make seizures look more like seizures; it makes each class more internally consistent, attenuating variation that inflates within-class scatter without carrying label information, while leaving the structure that distinguishes the classes where it was.

This is a reassuring mechanism for a clinical pipeline. A stage that raised separability by pushing the centroids apart would be reshaping the evidence on which a downstream diagnosis rests, and would fail in the way hardest to detect, by making borderline windows look more decisively like whichever class it favoured; shrinking only $\swithin$ makes no such commitment about any individual window.

\subsection{Discussion}

Read together, the two readouts say something the denoising literature does not. What made the positive control useful, a reduction of within-class scatter in task-relevant features, is not a property a fidelity score measures: fidelity is computed per window against a reference waveform, and is blind to how a population of windows is arranged relative to the class structure of the task, which is precisely where the benefit was found. Worse, the two objectives can conflict. A denoiser maximizing agreement with a per-window reference is rewarded for preserving every fluctuation that reference contains, including the subject- and session-specific variation that inflates $\swithin$; the operation that helps the task gives that variation up. The five audited denoisers are not failing at their jobs; they are doing a job whose objective was never checked against the one the pipeline needed.

The resulting recommendation is small and cheap. Denoising proposed for a task pipeline should be reported with at least one downstream-task score under an otherwise fixed pipeline, and with the two-term readout of Eq.~\eqref{eq:d}, needing only class labels. The second distinguishes a stage that quiets nuisance from one that manufactures class structure, a distinction invisible in a fidelity score or a task score taken alone.

\para{Limitations.} The audit is deliberately narrow so that the preprocessing is the only free variable: one task, one corpus, one downstream architecture. Whether the ordering in Table~\ref{tab:audit} holds for other tasks, particularly non-pathological ones whose class structure differs, is open. We applied the probe to the raw signal and the positive control; running it for each audited denoiser would say whether their failures share a geometric signature. It also evaluates a fixed six-descriptor space chosen for clinical interpretability, so a stage could help through structure those descriptors do not express.

\section{Conclusion}

We audited five standard scalp-EEG denoising methods inside a fixed clinical seizure-detection pipeline and found that none improves on the unprocessed signal, the supervised ones degrading the threshold-free ranking outright. Because the fidelity objective these methods optimize cannot be validated on real scalp recordings, we formalized a label-only, classifier-free probe that reads a preprocessing stage by the separability geometry it induces. On a transform that does help, it attributed the entire gain to a roughly one-quarter reduction of within-class scatter, with class separation untouched. What a downstream model benefits from is thus variance suppression on task-relevant structure, which fidelity metrics neither measure nor reward.

\para{Ethics and generative AI usage.} All recordings come from public corpora, de-identified and released under their source institutions' ethical approval and data use agreements; the analysis decodes no private mental content. A large language model assistant was used only to improve the clarity of the writing and to assist with \LaTeX{} formatting; it generated no ideas, methods or results.

\bibliographystyle{IEEEtran}
\bibliography{references}

@book{nunez2006electric,
  title={Electric fields of the brain: the neurophysics of {EEG}},
  author={Nunez, Paul L and Srinivasan, Ramesh},
  year={2006},
  publisher={Oxford University Press}
}

@article{uriguen2015eeg,
  title={{EEG} artifact removal---state-of-the-art and guidelines},
  author={Urig{\"u}en, Jose Antonio and Garcia-Zapirain, Bego{\~n}a},
  journal={Journal of Neural Engineering},
  volume={12},
  number={3},
  pages={031001},
  year={2015},
  publisher={IOP Publishing}
}

@article{jung2000blind,
  title={Removing electroencephalographic artifacts by blind source separation},
  author={Jung, Tzyy-Ping and Makeig, Scott and Humphries, Colin and Lee, Te-Won and McKeown, Martin J and Iragui, Vicente and Sejnowski, Terrence J},
  journal={Psychophysiology},
  volume={37},
  number={2},
  pages={163--178},
  year={2000},
  publisher={Cambridge University Press}
}

@article{hyvarinen2000independent,
  title={Independent component analysis: algorithms and applications},
  author={Hyv{\"a}rinen, Aapo and Oja, Erkki},
  journal={Neural Networks},
  volume={13},
  number={4-5},
  pages={411--430},
  year={2000},
  publisher={Elsevier}
}

@article{delorme2007enhanced,
  title={Enhanced detection of artifacts in {EEG} data using higher-order statistics and independent component analysis},
  author={Delorme, Arnaud and Sejnowski, Terrence and Makeig, Scott},
  journal={NeuroImage},
  volume={34},
  number={4},
  pages={1443--1449},
  year={2007},
  publisher={Elsevier}
}

@article{donoho1994ideal,
  title={Ideal spatial adaptation by wavelet shrinkage},
  author={Donoho, David L and Johnstone, Iain M},
  journal={Biometrika},
  volume={81},
  number={3},
  pages={425--455},
  year={1994},
  publisher={Oxford University Press}
}

@article{mullen2015real,
  title={Real-time neuroimaging and cognitive monitoring using wearable dry {EEG}},
  author={Mullen, Tim R and Kothe, Christian AE and Chi, Yu Mike and Ojeda, Alejandro and Kerth, Trevor and Makeig, Scott and Jung, Tzyy-Ping and Cauwenberghs, Gert},
  journal={IEEE Transactions on Biomedical Engineering},
  volume={62},
  number={11},
  pages={2553--2567},
  year={2015},
  publisher={IEEE}
}

@article{zhang2021eegdenoisenet,
  title={{EEGdenoiseNet}: a benchmark dataset for deep learning solutions of {EEG} denoising},
  author={Zhang, Haoming and Zhao, Mingqi and Wei, Chen and Mantini, Dante and Li, Zherui and Liu, Quanying},
  journal={Journal of Neural Engineering},
  volume={18},
  number={5},
  pages={056057},
  year={2021},
  publisher={IOP Publishing}
}

@article{chuang2025art,
  title={Augmenting brain-computer interfaces with {ART}: An artifact removal transformer for reconstructing multichannel {EEG} signals},
  author={Chuang, Chun-Hsiang and Chang, Kong-Yi and Huang, Chih-Sheng and Bessas, Anne-Mei},
  journal={NeuroImage},
  volume={310},
  pages={121123},
  year={2025},
  publisher={Elsevier}
}

@article{huang2025eegdfus,
  author={Huang, Xiaoyang and Li, Chang and Liu, Aiping and Qian, Ruobing and Chen, Xun},
  journal={IEEE Journal of Biomedical and Health Informatics}, 
  title={{EEGDfus}: A Conditional Diffusion Model for Fine-Grained {EEG} Denoising}, 
  year={2025},
  volume={29},
  number={4},
  pages={2557-2569},
  doi={10.1109/JBHI.2024.3504716}}

@article{choi2025targetedadversarialdenoisingautoencoders,
  title={Targeted adversarial denoising autoencoders ({TADA}) for neural time series filtration},
  author={Choi, Benjamin J and Milsap, Griffin and Scholl, Clara A and Tenore, Francesco and Ogg, Mattson},
  journal={arXiv preprint arXiv:2501.04967},
  year={2025}
}

@article{kayser2015surface,
  title={Issues and considerations for using the scalp surface {L}aplacian in {EEG/ERP} research: A tutorial review},
  author={Kayser, J{\"u}rgen and Tenke, Craig E},
  journal={International Journal of Psychophysiology},
  volume={97},
  number={3},
  pages={189--209},
  year={2015},
  publisher={Elsevier}
}

@inproceedings{esteller2001line,
  title={Line length: an efficient feature for seizure onset detection},
  author={Esteller, Rosana and Echauz, Javier and Tcheng, T and Litt, Brian and Pless, Benjamin},
  booktitle={Proceedings of the 23rd Annual International Conference of the IEEE Engineering in Medicine and Biology Society},
  volume={2},
  pages={1707--1710},
  year={2001},
  organization={IEEE}
}

@article{hjorth1970eeg,
  title={{EEG} analysis based on time domain properties},
  author={Hjorth, Bo},
  journal={Electroencephalography and Clinical Neurophysiology},
  volume={29},
  number={3},
  pages={306--310},
  year={1970},
  publisher={Elsevier}
}

@article{alam2013detection,
  title={Detection of seizure and epilepsy using higher order statistics in the {EMD} domain},
  author={Alam, SM Shafiul and Bhuiyan, Mohammed Imamul Hassan},
  journal={IEEE Journal of Biomedical and Health Informatics},
  volume={17},
  number={2},
  pages={312--318},
  year={2013},
  publisher={IEEE}
}

@article{pyrzowski2021zero,
  title={Zero-crossing patterns reveal subtle epileptiform discharges in the scalp {EEG}},
  author={Pyrzowski, Jan and Le Douget, Jean-Eudes and Fouad, Amal and Siemi{\'n}ski, Mariusz and J{\k{e}}drzejczak, Joanna and Le Van Quyen, Michel},
  journal={Scientific Reports},
  volume={11},
  number={1},
  pages={4128},
  year={2021},
  publisher={Nature Publishing Group UK London}
}

@book{cohen1988statistical,
  title={Statistical Power Analysis for the Behavioral Sciences},
  author={Cohen, Jacob},
  edition={2},
  year={1988},
  publisher={Lawrence Erlbaum Associates}
}

@book{green1966signal,
  title={Signal detection theory and psychophysics},
  author={Green, David Marvin and Swets, John A},
  year={1966},
  publisher={Wiley}
}

@phdthesis{shoeb2010chbmit,
  title={Application of machine learning to epileptic seizure onset detection and treatment},
  author={Shoeb, Ali Hossam},
  year={2009},
  school={Massachusetts Institute of Technology}
}

@article{thuwajit2022eegwavenet,
  author={Thuwajit, Punnawish and Rangpong, Phurin and Sawangjai, Phattarapong and Autthasan, Phairot and Chaisaen, Rattanaphon and Banluesombatkul, Nannapas and Boonchit, Puttaranun and Tatsaringkansakul, Nattasate and Sudhawiyangkul, Thapanun and Wilaiprasitporn, Theerawit},
  journal={IEEE Transactions on Industrial Informatics}, 
  title={{EEGWaveNet}: Multiscale {CNN}-Based Spatiotemporal Feature Extraction for {EEG} Seizure Detection}, 
  year={2022},
  volume={18},
  number={8},
  pages={5547-5557},
  doi={10.1109/TII.2021.3133307}}

@article{defossez2022encodec,
  title={High fidelity neural audio compression},
  author={D{\'e}fossez, Alexandre and Copet, Jade and Synnaeve, Gabriel and Adi, Yossi},
  journal={arXiv preprint arXiv:2210.13438},
  year={2022}
}

@inproceedings{carzaniga2025braincodec,
title={The Case for Cleaner Biosignals: High-fidelity Neural Compressor Enables Transfer from Cleaner i{EEG} to Noisier {EEG}},
author={Francesco S. Carzaniga and Gary Tom Hoppeler and Michael Hersche and Kaspar Schindler and Abbas Rahimi},
booktitle={The Thirteenth International Conference on Learning Representations},
year={2025},
url={https://openreview.net/forum?id=b57IG6N20B}
}

@inproceedings{burrello2019swec,
  title={Laelaps: An energy-efficient seizure detection algorithm from long-term human {iEEG} recordings without false alarms},
  author={Burrello, Alessio and Cavigelli, Lukas and Schindler, Kaspar and Benini, Luca and Rahimi, Abbas},
  booktitle={2019 Design, Automation \& Test in Europe Conference \& Exhibition (DATE)},
  pages={752--757},
  year={2019},
  organization={IEEE}
}

@article{ball2009signal,
  title={Signal quality of simultaneously recorded invasive and non-invasive {EEG}},
  author={Ball, Tonio and Kern, Markus and Mutschler, Isabella and Aertsen, Ad and Schulze-Bonhage, Andreas},
  journal={NeuroImage},
  volume={46},
  number={3},
  pages={708--716},
  year={2009},
  publisher={Elsevier}
}

@article{parvizi2018promises,
  title={Promises and limitations of human intracranial electroencephalography},
  author={Parvizi, Josef and Kastner, Sabine},
  journal={Nature Neuroscience},
  volume={21},
  number={4},
  pages={474--483},
  year={2018},
  publisher={Nature Publishing Group US New York}
}

@article{bao2026ieeg,
  title={{iEEG} Can Help {EEG}: Improving Clinical Scalp {EEG} Analysis through Intracranial Priors},
  author={Bao, Xianglong and Ding, Yi and Li, Zengxiang and Liu, Nan and Wang, Kewen and Wang, Zhe and Jin, Ming},
  journal={arXiv preprint arXiv:2611.20435},
  note={Accepted at KDD 2027},
  year={2026}
}

\end{document}